\documentclass[twoside,journey]{IEEEtran}
\usepackage{makecell}
\usepackage{hyperref}
\usepackage{array}
\usepackage{graphicx,amssymb,amsmath}
\usepackage{multicol}
\usepackage[noadjust]{cite}
\usepackage{setspace}
\usepackage{subfigure}
\usepackage{graphicx}
\usepackage{float}
\usepackage {url}
\usepackage{stfloats}
\usepackage{amsthm,pifont}
\usepackage{flushend}
\usepackage{cases,subeqnarray}
\usepackage{bm,multirow,bigstrut}
\usepackage{amsmath, amsthm, amssymb}
\usepackage{textcomp}
\usepackage{latexsym,bm}
\usepackage{booktabs}
\usepackage{xcolor}
\usepackage{mathtools}
\usepackage{dsfont}
\usepackage{extarrows}
\usepackage{epsfig}
\usepackage{epsfig}
\usepackage{epstopdf}
\usepackage[noend]{algpseudocode}
\usepackage{algorithmicx,algorithm}
\usepackage{svg}
\theoremstyle{plain}

\theoremstyle{plain}

\usepackage{amsmath}

\IEEEoverridecommandlockouts

\DeclareUnicodeCharacter{3000}{~}

\begin{document}

\title{Enabling AI-Generated Content (AIGC) Services in Wireless Edge Networks}
\author{Hongyang~Du, Zonghang Li, Dusit~Niyato,~\IEEEmembership{Fellow,~IEEE}, Jiawen~Kang, Zehui Xiong, Xuemin~(Sherman)~Shen,~\IEEEmembership{Fellow,~IEEE}, and Dong~In~Kim,~\IEEEmembership{Fellow,~IEEE}
\thanks{H.~Du, and D. Niyato are with the School of Computer Science and Engineering, the Energy Research Institute @ NTU, Interdisciplinary Graduate Program, Nanyang Technological University, Singapore (e-mail: hongyang001@e.ntu.edu.sg, dniyato@ntu.edu.sg).}
\thanks{Z. Li is with the School of Information and Communication Engineering,
University of Electronic Sciences and Technology of China, Chengdu,
China. (Email: lizhuestc@gmail.com).}
\thanks{J. Kang is with the School of Automation, Guangdong University of Technology, China. (e-mail: kavinkang@gdut.edu.cn)}
\thanks{Z. Xiong is with the Pillar of Information Systems Technology and Design, Singapore University of Technology and Design, Singapore (e-mail: zehui\_xiong@sutd.edu.sg)}
\thanks{X. Shen is with the Department of Electrical and Computer Engineering, University of Waterloo, Canada (e-mail: sshen@uwaterloo.ca)}
\thanks{D. I. Kim is with the Department of Electrical and Computer Engineering, Sungkyunkwan University, South Korea (e-mail: dikim@skku.ac.kr)}
}
\maketitle
\begin{abstract}
Artificial Intelligence-Generated Content (AIGC) refers to the use of AI to automate the information creation process while fulfilling the personalized requirements of users. However, due to the instability of AIGC models, e.g., the stochastic nature of diffusion models, the quality and accuracy of the generated content can vary significantly. In wireless edge networks, the transmission of incorrectly generated content may unnecessarily consume network resources. Thus, a dynamic AIGC service provider (ASP) selection scheme is required to enable users to connect to the most suited ASP, improving the users' satisfaction and quality of generated content. In this article, we first review the AIGC techniques and their applications in wireless networks. We then present the AIGC-as-a-service (AaaS) concept and discuss the challenges in deploying AaaS at the edge networks. Yet, it is essential to have performance metrics to evaluate the accuracy of AIGC services. Thus, we introduce several image-based perceived quality evaluation metrics. Then, we propose a general and effective model to illustrate the relationship between computational resources and user-perceived quality evaluation metrics. To achieve efficient AaaS and maximize the quality of generated content in wireless edge networks, we propose a deep reinforcement learning-enabled algorithm for the optimal ASP selection. Simulation results show that the proposed algorithm can provide a higher quality of generated content to users and achieve fewer crashed tasks by comparing with four benchmarks, i.e., overloading-avoidance, random, round-robin policies, and the upper-bound schemes.
\end{abstract}
\IEEEpeerreviewmaketitle
\section{Introduction}
Artificial Intelligence-Generated Content (AIGC) techniques have gained significant attention due to the unprecedented ability to automate the creation of various content~\cite{yunjiu2022artificial}, e.g., text, images, and videos. Undoubtedly, AIGC will significantly impact daily applications, especially Metaverse. With the ability to produce efficiently large amounts of high-quality content, AIGC can save time and resources that would otherwise be spent on manual content creation.

Recent research studies demonstrate that significant progress has been made in AIGC. Specifically, in text generation, the authors in \cite{chen2020generative} and \cite{guo2018long} have explored methods for generating coherent and diverse texts using deep learning techniques. For image generation, studies such as \cite{karras2018progressive} and \cite{huang2018multimodal} have focused on generating photo-realistic images using generative adversarial networks (GANs). In the audio generation, the authors in \cite{ping2020waveflow} have explored deep learning techniques for synthesizing high-quality speech. Furthermore, the diffusion model brings the latest breakthrough in the AIGC area. In 2020, GPT-3 model was published by OpenAI as a multimodal do-it-all language model that is capable of machine translation, text generation, semantic analysis, etc~\cite{floridi2020gpt}. Then, the diffusion model-based DALL-E2 released in 2022 is regarded as the state-of-the-art image generation model which can outperform GANs~\cite{dhariwal2021diffusion}.

However, AIGC models require a large amount of data for training, and the big AIGC models are difficult to be deployed. Taking Stable Diffusion for example, {\it{Stability AI}} company maintains over 4,000 NVIDIA A100 GPU clusters and has spent over \$50 million in operating costs (https://stability.ai/). The Stable Diffusion V1 requires 150,000 A100 GPU hours for a single training session. Moreover, AIGC models that are trained by different datasets are suitable for different tasks. For example, the AIGC model trained by the human face dataset can be used to repair corrupted face images, but may not be effective in correcting blurred landscape images. Due to the diversity of users' tasks and the limited edge device capacities, it is difficult to deploy multiple AIGC models on every network edge device. To further increase the availability of the AIGC services, one promising deployment scheme is based on ``Everything-as-a-service'' (EaaS), which can effectively provide users with subscription-based services. By embracing the EaaS deployment scheme, we present the concept of ``AIGC-as-a-service'' (AaaS). Specifically, AIGC service providers (ASPs) can deploy AI models on edge servers to deliver instant services to users over wireless networks, offering a more convenient and customizable experience. Users can easily access and enjoy AIGC with low latency and resource consumption. There are several advantages of deploying AaaS in edge networks:
\begin{enumerate}
\item[A1)] Personalization: AIGC models can be used to generate content tailored to each user's requirements, providing a personalized and engaging experience. For example, personalized product recommendations can be offered to users based on their locations, preferences, and usage patterns. 
\item[A2)] Efficiency: By deploying AIGC services closer to users, quality of services (QoS) will be improved significantly, e.g., lower delay, while network and computing resources can be utilized more efficiently due to local content transfer. 
\item[A3)] Flexibility: AIGC can be customized and optimized to meet dynamic demands and resource availability. By scheduling wireless network users' access for AIGC service providers, the overall QoS for users in the network can be maximized.
\end{enumerate}
Therefore, edge-based AaaS has the potential to revolutionize the way that content is created and delivered over wireless networks. However, the current research on AIGC focuses mainly on AIGC model training while ignoring the resource allocation issues when deploying AIGC in wireless edge networks. Specifically, AIGC may require significant bandwidth and computation power to generate and deliver content to users, which could lead to degraded network performance. Furthermore, scaling AaaS to meet the needs of a large number of users can be challenging. Thus, assigning suitable ASPs to users is critical. On the one hand, users pursue their goals of being served by the ASPs with the best performance. On the other hand, it is important to avoid overloading certain AIGC services and requiring re-transmissions, so as to consume scarce network resources. {\it{To the best of our knowledge, this is the first research work to discuss the deployments, aforementioned challenges, and future directions of AIGC in wireless edge networks.}} Our contributions can be summarized as follows:
\begin{itemize}
\item We provide a comprehensive overview of the AIGC and techniques behind it. Then, we discuss various applications of AIGC and their use cases in wireless edge networks and their deployment challenges.
\item We review the existing image-based perceived quality metrics. By conducting real experiments, we propose a general model to reveal the relationship between computational resource consumption and the quality of generated content in AaaS.
\item We propose a deep reinforcement learning (DRL)-enabled method to achieve a dynamic selection of optimal ASPs. We demonstrate the superiority of our proposed DRL-enabled algorithm compared with four solutions, including upper-bound, overloading-avoidance, random, and round-robin policies.
\end{itemize}

\section{AI-Generated Content and Techniques}
\begin{figure*}[h]
\centering
\includegraphics[width=0.9\textwidth]{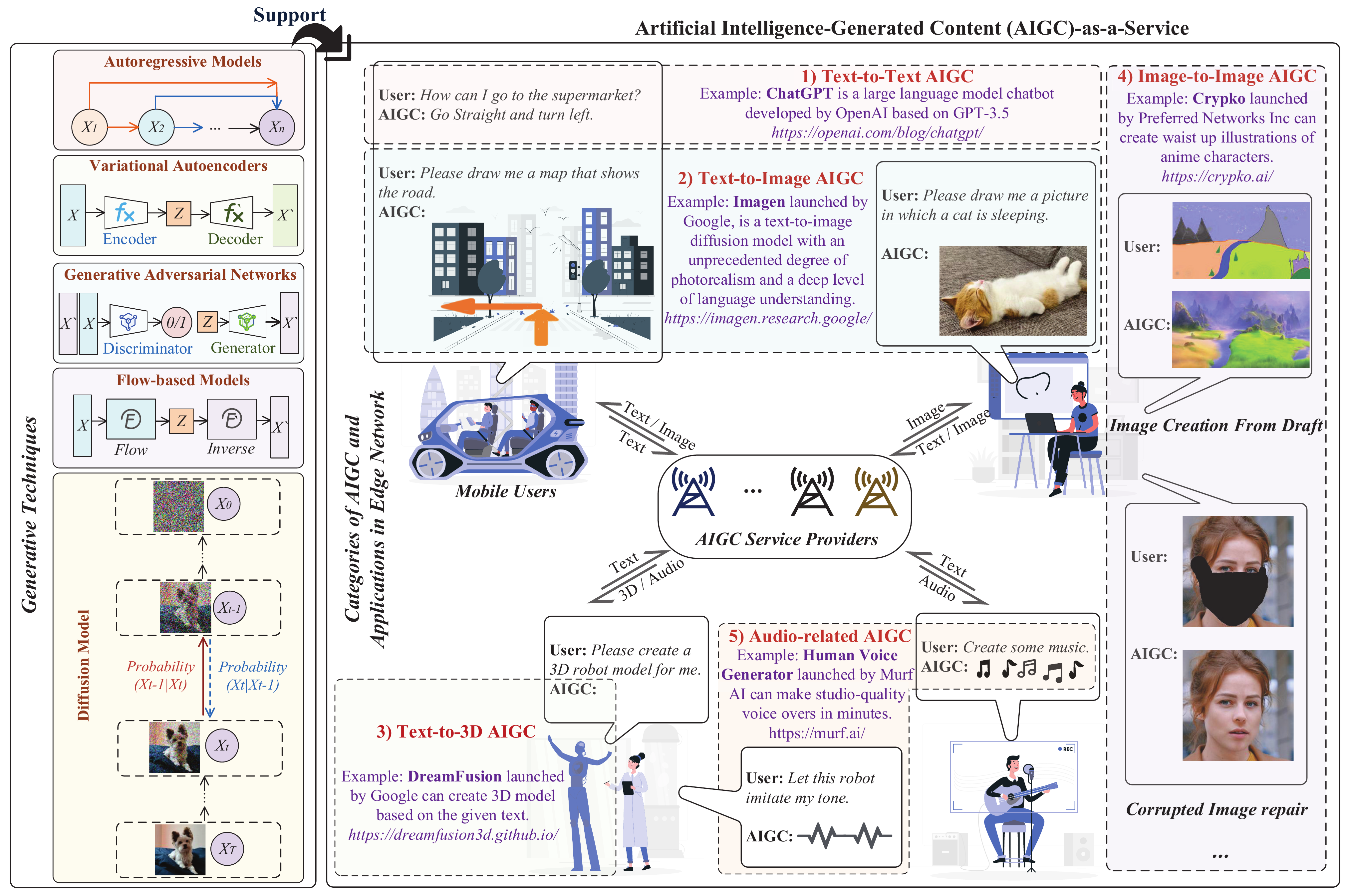}%
\caption{Generative techniques in AIGC~\cite{harshvardhan2020comprehensive}, categories of AIGC, and applications in wireless edge networks. We list several online available AIGC services as examples, e.g., ChatGPT for text-to-text AIGC (https://openai.com/blog/chatgpt/), Imagen for text-to-image AIGC (https://imagen.research.google/), DreamFusion for text-to-3D AIGC (https://dreamfusion3d.github.io/), Crypko for image-to-image AIGC (https://crypko.ai/), and Human Voice Generator for audio-related AIGC (https://murf.ai/).}
\label{framework}
\end{figure*}
In this section, we review the recent progress of AIGC. Specifically, we introduce the technologies behind the AIGC. Then, we discuss several categories of AIGC and associated applications in edge networks.

\subsection{Generative Techniques}
We introduce generative techniques in training AIGC models~\cite{harshvardhan2020comprehensive}. The basic model structures are shown on the left of Fig.~\ref{framework}.
\begin{itemize}
\item {\bf{Autoregressive Models (ARMs)}}: ARMs belong to statistical modeling that involves predicting the future values of a time series based on past values~\cite{harshvardhan2020comprehensive}. ARMs can generate text or other media types for content generation by predicting the next element based on the previous ones. A potential use case for ARMs is to generate music by predicting the next note in a musical sequence based on the previous notes from edge users.
\item {\bf{Variational Autoencoders (VAEs)}}: VAEs can generate new data by learning a compact, latent representation of the input data, consisting of an encoder network and a decoder network~\cite{harshvardhan2020comprehensive}. The encoder network processes the input data and outputs a latent representation. The decoder network takes this latent representation as input and generates synthetic data similar to the input data.
\item {\bf{Generative Adversarial Networks (GANs)}}: GANs consist of two neural networks, i.e., generator and discriminator networks \cite{karras2018progressive}. The two networks are trained together to improve the generator's ability to generate realistic images and the discriminator's ability to distinguish synthetic images from real images.
\item {\bf{Flow-based Models (FBMs)}}: FBMs transform a simple distribution into a target distribution through a series of invertible transformations~\cite{harshvardhan2020comprehensive}. These transformations are implemented as neural networks, and the process of applying the transformations is referred to as ``flow''.
\item {\bf{Diffusion Models (DMs)}}: DMs are trained to denoise images blurred by Gaussian noise to learn how to reverse the diffusion process~\cite{dhariwal2021diffusion}. Several diffusion-based generative models have been proposed, including diffusion probabilistic models, noise-conditioned score networks, and denoising diffusion probabilistic models.
\end{itemize}
Moreover, classic techniques such as Transformer can also be used to train AIGC models, which are discussed in the following.

\subsection{Categories of AIGC and Applications in Mobile Networks}
We then present several categories of AIGC technologies and their applications in edge networks, which can serve as potential future research directions.
\subsubsection{Text-to-Text AIGC}
Text-to-text AIGC can generate the human-like message as an output based on a given text input. Therefore, it can be used for automatic answers, language translation, or article summarization.
One representative text-to-text AIGC model is the Generative Pre-training Transformer (GPT) ({\it{https://openai.com/blog/chatgpt/}}), a language model developed by OpenAI~\cite{floridi2020gpt}. The GPT is trained on a large dataset of human-generated text, such as books or articles. The model can then create text by predicting the next word in a sequence based on the words that come before it. GPT has been highly successful and has achieved state-of-the-art results on several natural language processing (NLP) benchmarks. GPT can be used to build many popular language-based services. In wireless edge networks, as shown in Fig.~\ref{framework}, GPT can serve as a chatbot that provides drivers with navigation and information alert services.
\subsubsection{Text-to-Image AIGC}
Text-to-image AIGC allows users to generate images based on text input, enabling the creation of visual content from written descriptions. It can be regarded as a combination of natural language processing and computer vision techniques. As shown in Fig.~\ref{framework}, the text-to-image AIGC can assist mobile users with various activities. For example, users in Internet-of-Vehicles can request visual-based path planning. Furthermore, text-to-image AIGC can also assist users in creating art and producing pictures in various styles based on users' descriptions or keywords.

\subsubsection{Text-to-3D AIGC}
Text-to-3D AIGC can generate 3D models from text descriptions while using wireless AR applications. Typically, generating 3D models requires higher computational resources than generating 2D images. Considering the development of next-generation Internet services, such as Metaverse~\cite{du2022attention}, generating 3D models based on text without complicated manual design is fascinating.

\subsubsection{Image-to-Image AIGC}
Image-to-Image AIGC uses AI models to generate realistic images from source images or create a stylized version of an input image. For example, when it comes to assisting artwork creation, image-to-image AIGC can generate visually satisfying pictures based solely on user-inputted sketches. Furthermore, image-to-image AIGC can be used for image editing services. Users can remove occlusions in one image or repair corrupted images.

\subsubsection{Audio-related AIGC}
Audio-related AIGC models analyze, classify, and manipulate audio signals, including speech and music. Specifically, text-to-speech models are designed to synthesize natural-sounding speech from text input. Music generation models can synthesize music in a variety of styles and genres. Audio-visual music generation involves using both audio and visual information, such as music videos or album artwork, to generate music compositions that are more closely tied to a particular visual style or theme. Moreover, audio-related AIGC can serve as voice assistants that answer users' queries. Alexa (https://developer.amazon.com/en-US/alexa) and Siri (https://www.apple.com/sg/siri/) are examples of real-life applications.

Given the power of AIGC models, there are several challenges in deploying AaaS in wireless edge networks, which are introduced in the following.

\section{AI-Generated Content-as-a-Service in Wireless Edge Networks}
In this section, we discuss the AaaS in detail, including the challenges and performance metrics.
\subsection{AI-Generated Content-as-a-Service and Challenges}
To deploy AaaS in wireless edge networks, the ASPs should first train AIGC models on large datasets. The AIGC models would need to be hosted on edge servers and can be accessed by users. Continuous maintenance and updates would be required to ensure that the AIGC models remain accurate and effective for generating high-quality content. Users can submit requests for content generation and receive the generated content from edge servers rented by ASPs. Despite several advantages of deploying AaaS in wireless edge networks, there are pertaining challenges to be addressed. 


\begin{itemize}
\item {\bf{Bandwidth Consumption}}: The AIGC consumes a significant amount of bandwidth. Especially for AaaS related to high-resolution images, both upload and download processes require considerable network resources to ensure low-latency services. For example, the data size of an AI-generated wallpaper in {\it{wallhaven (https://wallhaven.cc/tag/133451)}} can be around $10$ Megabytes. Furthermore, due to the diversity of the generated images, users may make multiple repeated requests to specific edge servers to obtain a satisfactory image.
\item {\bf{Time-varying Channel Quality}}: The QoS in AaaS can be affected by the wireless transmission of the generated content. Low Signal-to-Noise Ratio (SNR), low Outage Probability (OP), and high bit-error probability (BEP) can degrade QoS of AIGC services and users' satisfaction, which results from time-varying fading channels when the channel encounters deep fading occasionally.

\item {\bf{Dataset used for training AIGC Models}}: The dataset used for training AIGC models can impact the quality of the generated content. Since different ASPs have various AIGC models, users can be allocated to the suitable ASP to meet their requirements. For example, AIGC models trained with more face images will be more suitable for generating avatars than those trained with other datasets.
\item {\bf{Computation Resource Consumption}}: The well-trained AIGC model still consumes time and computational resources when generating content, e.g., fine-tuning and inference. For example, the quality of the output of the diffusion model-AaaS increases with the number of inference steps.
\item {\bf{Utility Maximization and Incentive Mechanism}}: Incentive mechanism design is significant in AaaS as it can motivate ASPs to generate high quality content, meeting the desired goals and objectives. Here, the utility function should include the perceived QoS from users.
\end{itemize}
\begin{figure*}[h]
\centering
\includegraphics[width=0.98\textwidth]{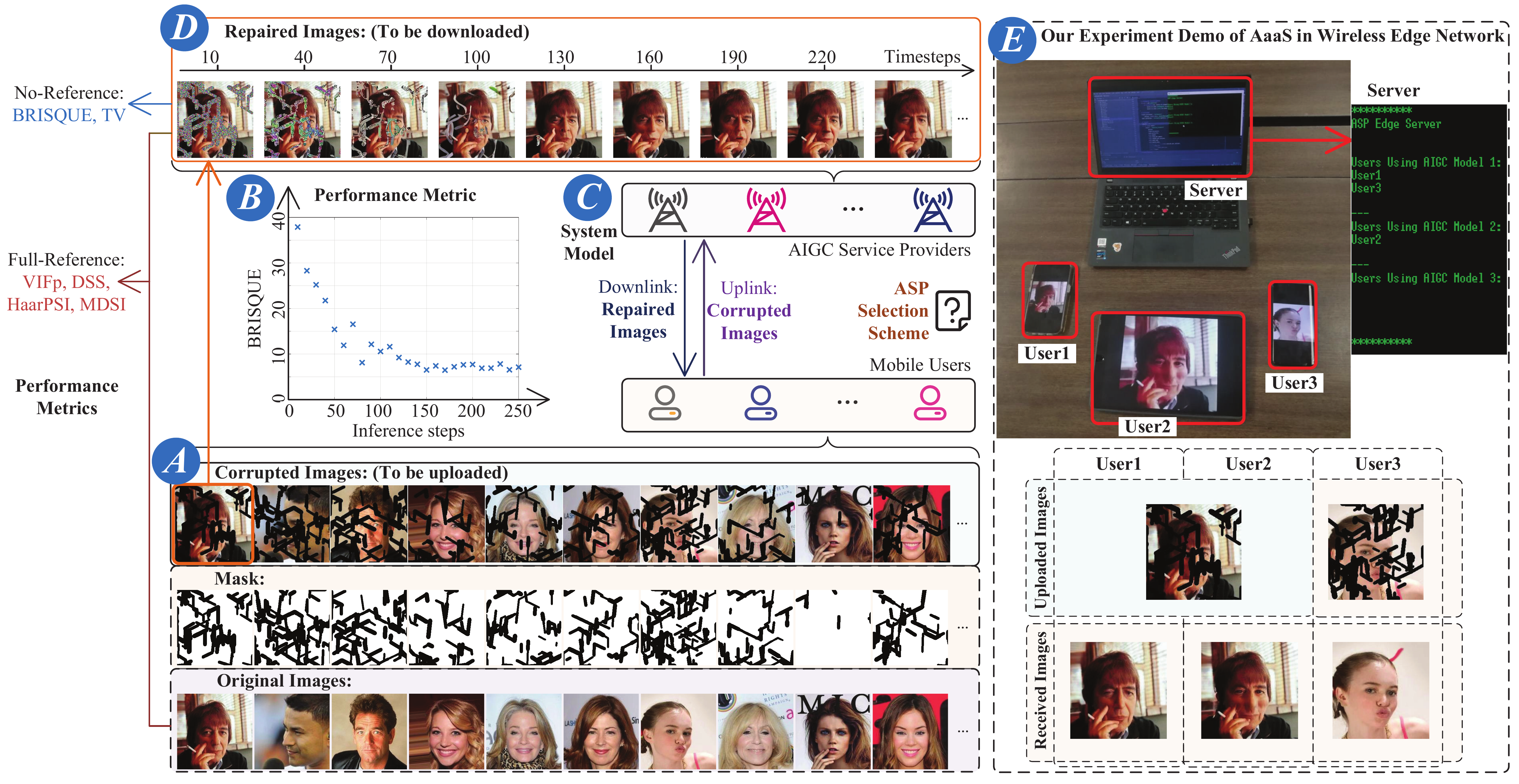}%
\caption{Example of an AaaS for repairing corrupted images. The corrupted images are shown in Part A, and the repaired images under different inference steps are shown in Part D. Part B shows how the performance metric, BRISQUE, varies over different inference steps. Part C shows the system model of ASP selection problem. A experiment demo is shown in Part E.}
\label{model1}
\end{figure*}
A common issue in addressing the above challenges is evaluating AIGC performance. Although many evaluation metrics in various modalities have been proposed, most of them are based on AI models or are difficult to be calculated, without a mathematical expression. For the optimal design of AaaS in wireless networks, AI-based resource allocation solutions can utilize AI-based performance metrics to consider the subjective feelings of the user. However, traditional mathematical resource allocation schemes require modeling the relationship between the computational resources consumption, e.g., the number of inference steps in the diffusion model, and the quality of the generated content, as shown in Fig.~\ref{model1}. To solve this problem, taking image-related AaaS as an example, we introduce various performance evaluation metrics and explore the mathematical relationship between metric values and computational resource consumption in the following.

\subsection{Performance Metric Modelling}\label{sec3c}
We first discuss AIGC evaluation metrics. We focus on evaluating the perceived quality of images, but the same methodology can be applied to other types of content. Then, we formulate the relationship between computational resource consumption and the quality of generated content in AaaS.

\subsubsection{Image-based metrics}
The image quality assessment metrics can be distribution-based and image-based. The distribution-based metrics, e.g., Frechet inception distance~\cite{piq}, take a list of image features to compute the distance between distributions for evaluating generated images. However, for practical AaaS in the wireless network, the quality evaluation is subjective, and it is hard for users to calculate distribution-based metrics. Thus, we focus on image-based metrics that attempt to achieve consistency in quality prediction by modeling salient physiological and psycho-visual features of the human visual system or by signal fidelity measures.

Specifically, without access to the original image as a reference, {\textit{no-reference image quality evaluation}} methods can be considered~\cite{piq}:
\begin{itemize}
\item {\bf{Total Variation (TV)}}: TV is a measure of the smoothness of an image. One common way to compute total variation is to take the sum of the absolute differences between adjacent samples in an image. This measures the "roughness" or "discontinuity" of the image.
\item {\bf{Blind/Referenceless Image Spatial Quality Evaluator (BRISQUE)}}\footnote{http://live.ece.utexas.edu/research/quality/}: BRISQUE utilizes scene statistics of locally normalized luminance coefficients to quantify possible losses of ``naturalness'' in the image due to distortions~\cite{mittal2012no}. It has been shown that BRISQUE performs well in correlation with human perception of quality.
\end{itemize}
The higher the image quality, the smaller the values of TV and BRISQUE.

For AaaS where a reference image is available, we can use {\textit{full-reference image quality evaluation}} methods \cite{piq}:
\begin{itemize}
\item {\bf{Discrete Cosine Transform Subbands Similarity (DSS)}}: DSS exploits essential characteristics of human visual perception by measuring changes in structural information in subbands in the discrete cosine transform (DCT) domain and weighting the quality estimates for these subbands~\cite{gatys2016neural}.
\item {\bf{Haar Wavelet-based Perceptual Similarity Index (HaarPSI)}}: HaarPSI utilizes the coefficients obtained from a Haar wavelet decomposition to assess local similarities between two images, as well as the relative importance of image areas.
\item {\bf{Mean Deviation Similarity Index (MDSI)}}: MDSI is a reliable and complete reference perceptual image quality assessment model that utilizes gradient similarity, chromaticity similarity, and deviation pooling.
\item {\bf{Visual Information Fidelity (VIF)}}: VIF is a competitive way of measuring fidelity that relates well with visual quality, which quantifies the information in the reference image and how much of the reference information can be extracted from the distorted image.
\end{itemize}
The higher the image quality, the higher the values of the aforementioned full-reference image quality metrics.

\subsubsection{A General Modelling of Perceived Image Quality Metric}
\begin{figure*}[h]
\centering
\includegraphics[width=0.9\textwidth]{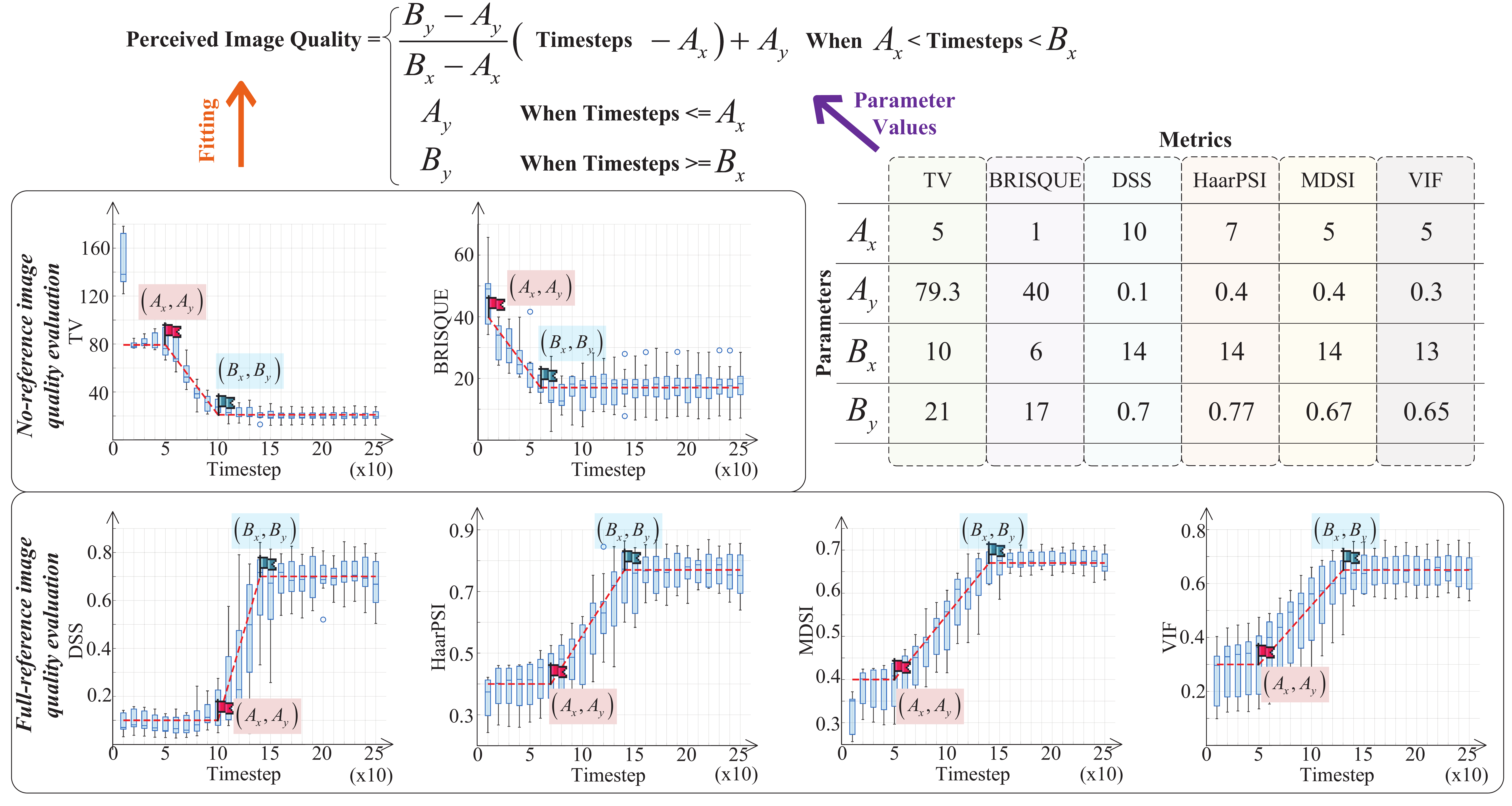}%
\caption{The relationship between the number of inference steps and various perceived image quality metrics, i.e., TV, BRISQUE, DSS, HaarPSI, MDSI, and VIF.}
\label{per}
\end{figure*}
AIGC models based on diffusion models are becoming mainstream. As shown in the left of Fig.~\ref{framework}, the diffusion process can be regarded as a step-wise denoising process. Thus, increasing the number of inference steps will improve the perceived image quality. However, the generated image quality does not always increase with the number of steps. Excessive inference steps incur unnecessary resource consumption. We conduct real experiments to investigate the relationship between the number of inference steps and various perceived image quality metrics, i.e., TV, BRISQUE, DSS, HaarPSI, MDSI, and VIF.

The experimental platform is built on a generic Ubuntu 20.04 system with an AMD Ryzen Threadripper PRO 3975WX 32-Cores CPU and an NVIDIA RTX A5000 GPU. We take diffusion model-based corrupted image restoration service as an example of AaaS. Specifically, we deploy the well-trained model, {\it{RePaint}}, proposed in \cite{lugmayr2022repaint} on our server. As shown in Fig.~\ref{model1} (Part A), we first generate a series of corrupted images, e.g., 20 images, with the help of masks. Then, these corrupted images are fed into {\it{RePaint}}. We can observe that the corrupted image gradually recovers as the inference progresses, as shown in Fig.~\ref{model1} (Part D). Moreover, the values of image quality metric, e.g., BRISQUE, decrease, as shown in Fig.~\ref{model1} (Part B). We show the values of each performance metric under the different number of timesteps in Fig.~\ref{per}.

Thus, we present a general model of the perceived image quality metric that contains four parameters, as shown at the top of Fig.~\ref{per}. Specifically, $A_x$ is the minimum number of inference steps where the image quality starts to improve, $A_y$ is the lower bound of the image quality, which can be regarded as the evaluation value for images with high noise, $B_x$ is the number of inference steps when the image quality starts to stabilise because of the capability of AIGC models, and $B_y$ is the highest image quality value that the model can achieve. Regardless of whether the performance metric value is positively or inversely proportional to the image quality, and regardless of the AaaS types, we can easily find the points $\left(A_x, A_y\right)$ and $\left(B_x, B_y\right)$ experimentally, as shown in Fig.~\ref{per}. 

{\bf{Lesson Learned:}} {\it{Despite the inherent uncertainty of the diffusion process, from Fig.~\ref{per}, we can observe that the perceived image quality increases or decreases approximately proportionally with the increase of inference steps. In the practical AIGC model analysis, we can perform experiments with the simple fitting method as shown in Fig.~\ref{per} to a performance metric to obtain four parameters in our proposed general mathematical model. Then, the model can be used in wireless edge network-enabled AIGC services analysis.}}

\section{Deep Reinforcement Learning-aided Dynamic ASP Selection}\label{sec4}

In this section, we study the optimal ASP edge server selection problem. We propose a DRL-enabled solution to maximize utility function while satisfying users' requirements.

\subsection{AaaS System Model}

Our demo is shown in Fig.~\ref{model1} (Part E). Specifically, three users are selecting between two image reparation AIGC models that are trained on datasets CelebA-HQ and Places2~\cite{lugmayr2022repaint}, respectively. User 1 and User 2 upload the same corrupted images. We can observe that different AIGC models will create different results for the same user task.

Then, we study the case for large-scale deployment of AaaS in wireless edge networks. We consider 20 AIGC service providers (ASPs) and 1000 edge users in the simulation. Each ASP provides AaaS with maximal resource capacity, i.e., total diffusion timesteps within a time window, ranging from 600 to 1500 at random. Each user submits multiple AIGC task requests to ASPs at different times. These tasks specify the amount of AIGC resources that they need, i.e., diffusion timesteps, which we set as a random value between 100 and 250. The user task arrivals follow the Poisson distribution. Specifically, during a period of 288 hours, user tasks arrive at the rate of $\lambda=0.288$ hour/request and there are a total of 1000 tasks. Note that the quality of AIGC models provided by different ASPs is different, e.g., the repaired images could be more realistic and natural.

A simple, yet less effective ASP selection, is that the user sends the task request directly to the ASP with the best quality of generated content. However, this approach will inevitably overload some ASPs due to insufficient computational resources and interrupting tasks in practice. In addition, the quality of generated content of ASPs is unknown to users. Mobile users need to ask ASPs several times to estimate the quality of generated content to execute myopic selection, which introduces unnecessary load and wireless network resource consumption. To this end, under the premise of the unknown quality of generated content, how to choose a suitable ASP for user tasks to maximize the overall system's utility and reduce AIGC resource overload and interruption caused by popular ASPs, is a challenging yet important problem.


\subsection{Deep Reinforcement Learning-based Solution}
\begin{figure}[h]
\centering
\includegraphics[width=0.45\textwidth]{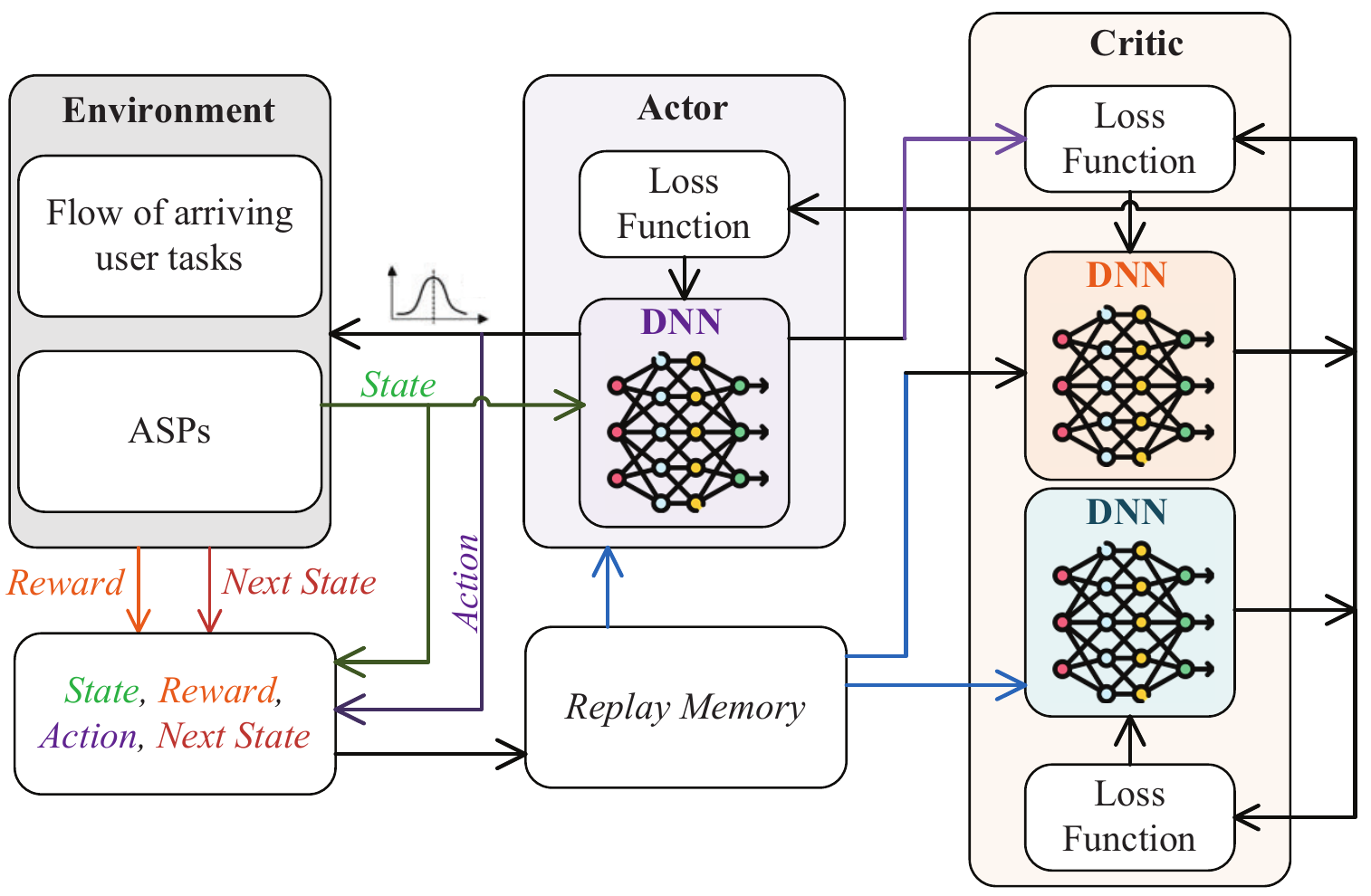}%
\caption{The structure of soft actor–critic DRL algorithm.}
\label{sdrl}
\end{figure}
We use the soft actor–critic (SAC) DRL~\cite{christodoulou2019soft} to solve the above dynamic ASP selection problem. As shown in Fig.~\ref{sdrl}, the learning process alternates
between policy evaluation (Critic) and policy improvement (Actor). Unlike the conventional actor-critic architecture, the policy in SAC is trained to maximize a trade-off between the expected return and entropy. The state space, action space, and reward in the AaaS environment are defined as follows:
\begin{itemize}
    \item \textbf{State:} The state space is composed of two parts: (a) a feature vector ({\it{the demand of AIGC resources for current user task and the estimated completion time of the task}}) of the arriving user task, and (b) feature vectors ({\it{the total AIGC resources of the $i$-th ASP and the currently available resources of the $i$-th ASP}}) of all ASPs in the current state.
    \item \textbf{Action.} The action space of the ASP selection problem is an integer indicating the selected ASP. In detail, the actor policy network outputs a 20-dimensional logits vector, and then the probability of selecting each ASP is obtained after being post-processed by the softmax operator. Finally, DRL selects an ASP to handle the current user task according to the assigned probability of each ASP.
    \item \textbf{Reward.} The reward consists of two parts: a quality of generated content reward and an congestion penalty. The former is defined as the perceived quality of the repaired image, as discussed in Section~\ref{sec3c}. Furthermore, any action that overloads AIGC models must be penalized as an congestion penalty. First, the action itself should be punished with fixed {\it{penalty value}}. Second, considering that ill-considered actions can cause bottleneck ASP's model to crash and the running tasks will be interrupted, the current action will also be subject to additional penalties according to the progress of ongoing tasks. The total reward returned is the quality reward minus the congestion penalty. Note that a larger {\it{penalty value}} will encourage DRL to pay more attention to avoid crashes.
\end{itemize}

We compare the performance of the DRL-enabled ASP selection algorithm with four benchmarks. The lower bound is the random allocation policy, assigning every new user task to an ASP randomly. In contrast, the optimal policy gives an approximate upper-bound on the performance, assuming that the quality scores available for each task on all ASPs are known (which is a posterior knowledge and is rarely satisfied in practice). The upper-bound policy can use the greedy algorithm to allocate a new user task to the ASP with enough AIGC resources and the highest quality. Furthermore, we implement the round-robin and overloading-avoidance policies, which are widely adopted in web applications to realize load balancing. It is simple, easy to implement, and starvation-free. The overloading-avoidance policy assigns the new user task to the ASP with most AIGC resources currently available to prevent or reduce the severity of overloads and crashes.

\begin{figure}[t]
\centering
\includegraphics[width=0.5\textwidth]{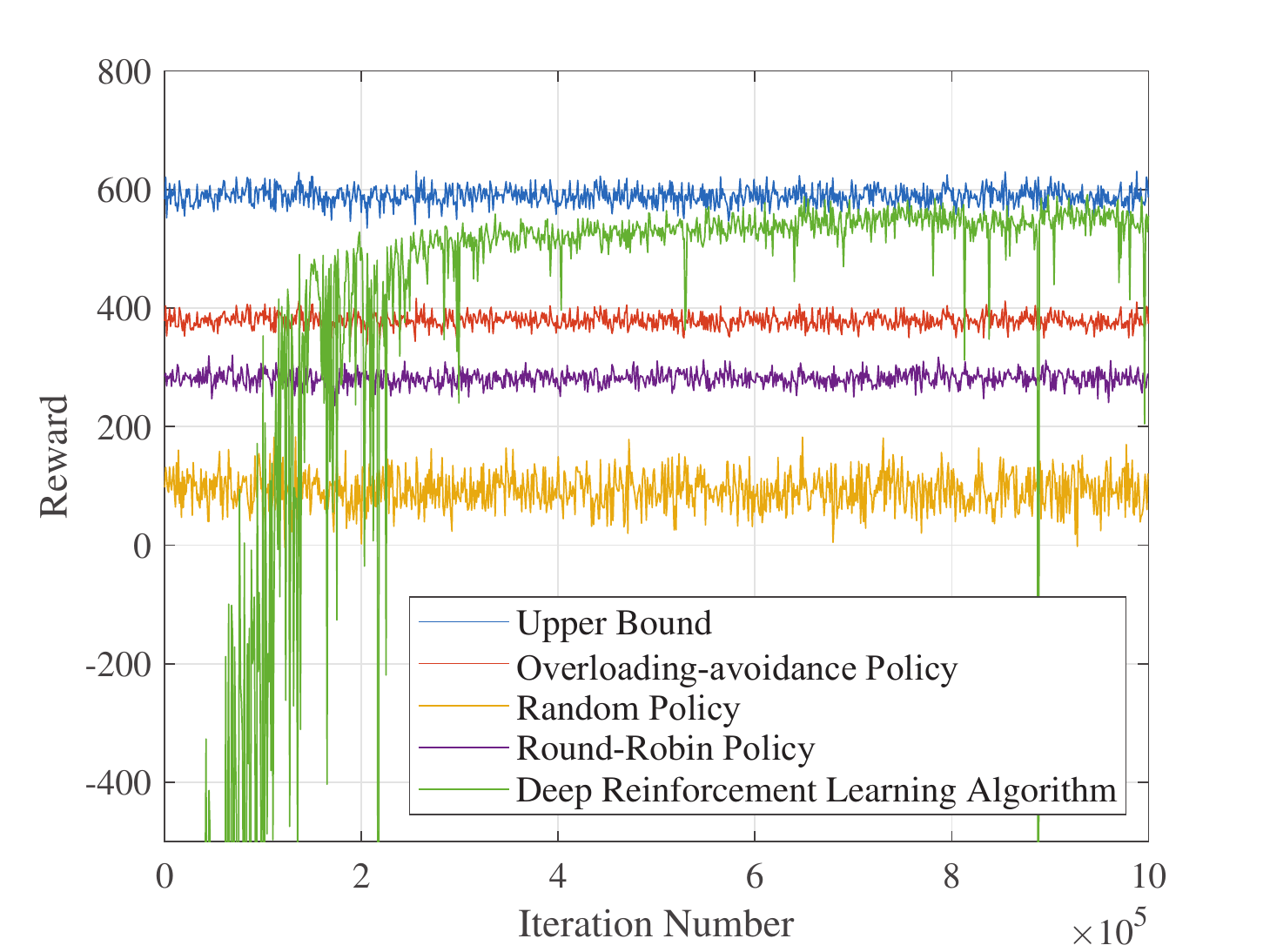}%
\caption{Reward values versus the number of iteration number in DRL. For performance comparison, we show the results of overloading-avoidance, random, round-robin policies, and their upper bound.}
\label{perdrl}
\end{figure}

\begin{figure}[t]
\centering
\includegraphics[width=0.5\textwidth]{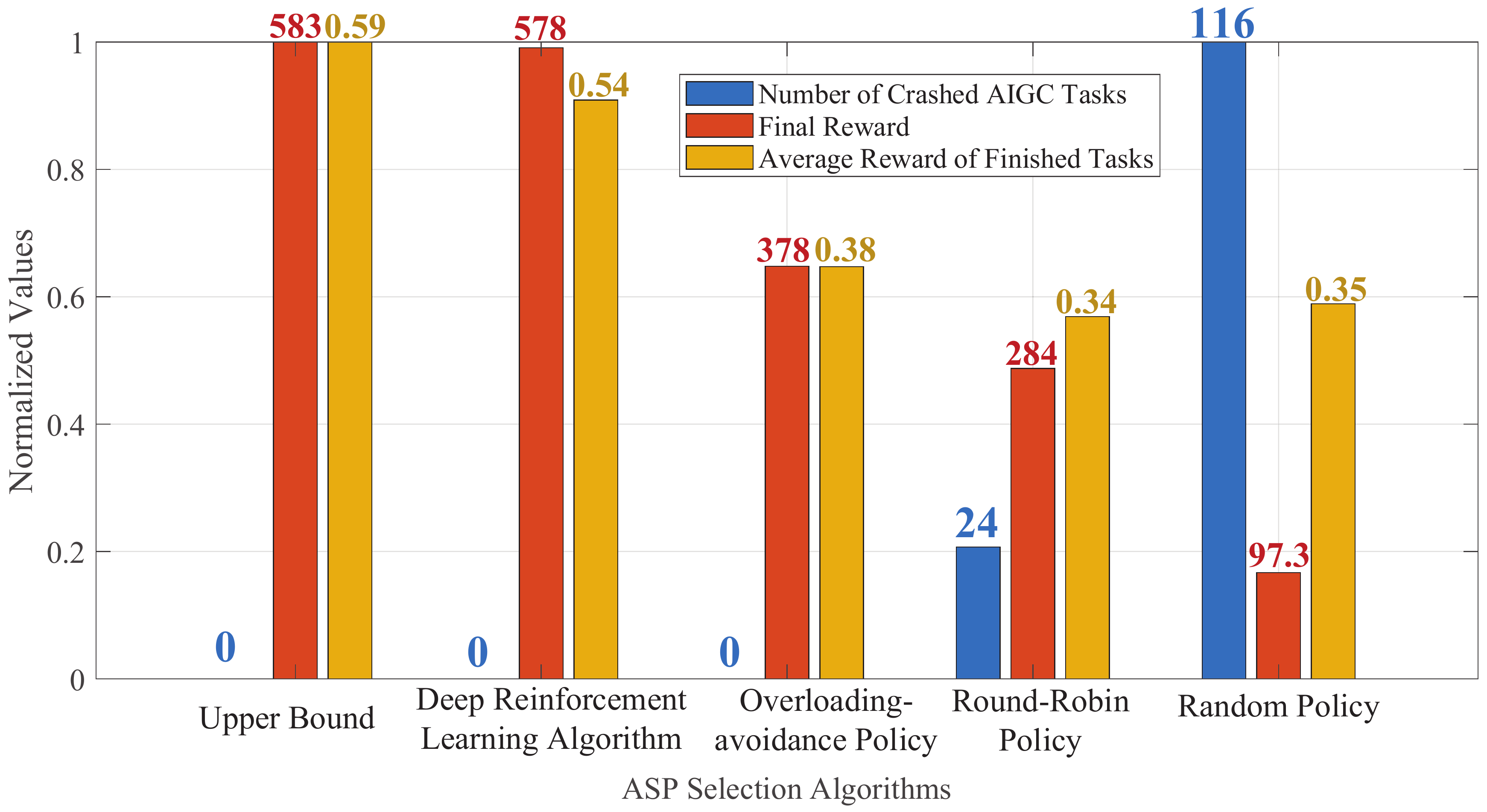}%
\caption{Comparison of episodic rewards, average rewards of finished tasks, and number of crashed tasks.}
\label{bar-fig}
\end{figure}

Figure~\ref{perdrl} shows the utility curves (i.e., reward curves) of the DRL-enabled ASP selection policy and the four benchmark policies. Since DRL can learn and evolve, as the learning step progresses, DRL has a more comprehensive and accurate selection of the ASP. Thus the utility rises rapidly, showing unique learning ability. One interesting finding is that when DRL overtakes the round-robin, DRL already has a specific load-balancing capability. Immediately afterward, DRL surpasses overloading-avoidance. At this time, DRL has learned to avoid actions that may cause crashes, thereby avoiding the congestion penalty. Then, DRL starts to learn the priority of different ASPs, and it seeks to place the current user task on the ASP with high quality to maximize the reward. Therefore, as shown in Fig.~\ref{perdrl}, DRL still has much room for improvement after surpassing the overloading-avoidance policy and finally reaching an episodic reward comparable to the upper-bound policy.

Figure \ref{bar-fig} counts the episodic rewards, the average rewards of finished tasks, and the number of crashed tasks of the five policies. On the one hand, the DRL-enabled ASP selection policy achieves zero task crashes and minimizes the congestion penalty, which is critical to providing a satisfying quality of generated content to users. On the other hand, DRL policy can learn the quality of content that ASPs may provide, which is unknown in other policies. Then, DRL can assign user tasks to ASPs that can provide higher QoS so that the average reward for each task is effectively increased. The combination of the above two advantages makes a much higher cumulative episodic reward for the DRL-enable ASP selection policy.

\section{Future Direction}
\subsection{Secure AIGC-as-a-Service}
When deploying AaaS in a wireless network, the requests from users and the generated contents from ASPs are transmitted in a wireless environment. Therefore, advanced security techniques for AIGC need to be studied, e.g., protecting the transmission of AIGC data through improved physical layer security techniques. Moreover, blockchain can be used to enable decentralized content distribution, allowing content to be shared and accessed directly between users without the need for a central authority. The authenticity and provenance of AIGC can be verified with the aid of blockchain, helping to ensure that the AIGC is accurate and trustworthy. Furthermore, during the training process of AIGC models, the privacy of the training data needs to be guaranteed, especially biometric data, such as face images. One possible solution is to train the model by secure federated learning.

\subsection{IoT-based and Wireless Sensing-aided Passive AaaS}
 Considering the fast development of sensing technologies, we aim to enable ubiquitous passive AaaS with wireless sensing signals. For example, wireless sensors can gather data about the environment or user behavior, which can then be fed into an AIGC model to generate relevant content. 
 Wireless sensing-aided passive AaaS can also be used in healthcare. Specifically, by using IoT devices to detect users' activity levels, sleep patterns, or heart rates with the aid of wireless sensing, the AIGC could generate content such as personalized workout plans. Moreover, the mobility of network devices predominantly affects the throughput of the connected links for AaaS, which is worth further study.

\subsection{Personalized Resource Allocation in AaaS}
Although the current AIGC models can meet users' needs with customized tasks, more studies are needed to achieve personalized AIGC services. For example, for text-to-image AaaS, when both users enter the text ``A monkey is standing next to a zebra'', current ASPs will produce similar images for users. However, if we deduce that the two users are a horse trainer and a monkey researcher, respectively, we can personalize the computing resource allocation~\cite{du2022attention}. Specifically, more computing resources should be allocated to generate and transmit the zebra in the image for the horse trainer. For the monkey researcher, the AIGC model that is more adapted to generating monkey images should be assigned to handle the task. One potential solution is incorporating user feedback and preferences into the content generation process and developing techniques for evaluating the effectiveness of personalized content.
\vspace{1cm}
\newpage

\section{Conclusion}\label{Con}
In this article, we reviewed the AIGC technologies and discussed the applications in wireless networks. To provide AIGC services to users, we proposed the concept of AaaS. Then, the challenges of deploying AaaS in wireless networks are discussed. In addressing these challenges, a fundamental problem is about the mathematical relationship between the resource consumption and the perceived quality of the generated content. After exploring various image-based performance evaluation metrics, we proposed a general modeling equation. Moreover, we studied the important ASP selection problem. A DRL-enabled algorithm was used to achieve nearly optimal ASP selection. As the first article to discuss AIGC in wireless networks, we hope that this article can inspire researchers to contribute to the advancement of wireless edge networks-enabled AaaS.

\bibliographystyle{IEEEtran}
\bibliography{Ref}
\end{document}